\documentclass[11pt]{article}

\usepackage[final]{acl}

\usepackage{times}
\usepackage{latexsym}

\usepackage[T1]{fontenc}
\usepackage[utf8]{inputenc}

\usepackage{microtype}

\usepackage{inconsolata}

\usepackage{graphicx}

\usepackage{booktabs} 
\usepackage{subcaption} 

\usepackage{paralist}
\usepackage[edges]{forest}
\usepackage{siunitx}
\usepackage{tikz-dependency}
\usepackage{tikz}
\usetikzlibrary{trees}
\usetikzlibrary{arrows.meta,shapes,positioning} 
\usepackage{forest} 
\usepackage{xcolor} 

\usepackage{cleveref}
\crefname{section}{§}{§}
\Crefname{section}{§}{§}

\title{A Survey of Inductive Reasoning for Large Language Models}

\author{
\textbf{Kedi Chen\textsuperscript{1,2}},
 \textbf{Dezhao Ruan\textsuperscript{1}},
 \textbf{Yuhao Dan\textsuperscript{1}},
 \textbf{Yaoting Wang\textsuperscript{3}},
 \textbf{Siyu Yan\textsuperscript{1}},
 \\
 \textbf{Xuecheng Wu\textsuperscript{4}},
 \textbf{Yinqi Zhang\textsuperscript{1}},
 \textbf{Qin Chen\textsuperscript{1}},
 \textbf{Jie Zhou\textsuperscript{1}},
 \textbf{Liang He\textsuperscript{1}},
 \\
 \textbf{Biqing Qi\textsuperscript{5}},
 \textbf{Linyang Li\textsuperscript{5}},
 \textbf{Qipeng Guo\textsuperscript{2,5}},
 \textbf{Xiaoming Shi\textsuperscript{1}},
 \textbf{Wei Zhang\textsuperscript{1,2}\thanks{Corresponding author.}}
\\
 \textsuperscript{1}East China Normal University,
 \textsuperscript{2}Shanghai Innovation Institute,
 \\
 \textsuperscript{3}Fudan University,
 \textsuperscript{4}Xi'an Jiaotong University,
 \textsuperscript{5}Shanghai AI Laboratory
 \\
 \texttt{\textbf{Email}}: \texttt{kdchen2@stu.ecnu.edu.cn}, \texttt{zhangwei.thu2011@gmail.com}
\\
\\
\textbf{Paper List}: \textcolor{blue}{https://github.com/BDML-lab/llm-inductive-reasoning-survey}
}

\begin{document}
\maketitle

\begin{abstract}
Reasoning is an important task for large language models (LLMs).
Among all the reasoning paradigms, inductive reasoning is one of the basic types,
which is characterized by its particular-to-general thinking process and the non-uniqueness of its answers.
The inductive mode is crucial for knowledge generalization and aligns better with human cognition, so it is a fundamental mode of learning, hence attracting increasing interest.
Despite the importance of inductive reasoning, there is no systematic summary of it. 
Therefore, this paper presents the first comprehensive survey of inductive reasoning for LLMs.
First, methods for improving inductive reasoning are categorized into three main areas: post-training enhancement, test-time exploration, and data augmentation. 
Then, current benchmarks of inductive reasoning are summarized, and a unified sandbox-based evaluation approach with the observation coverage metric is derived.
Finally, we offer some analyses regarding the source of inductive ability and how simple model architectures and data help with inductive tasks, providing a solid foundation for future research.

\end{abstract}

% by Xuecheng
% Large language models (LLMs) have made impressive progress in various downstream tasks, especially reasoning. Existing reasoning paradigms for LLMs can be primarily categorized into deductive reasoning and inductive reasoning. The former, represented by the mathematical and code reasoning, has already been extensively studied in recent years. The latter is characterized by its particular-to-general thinking process and the non-uniqueness of its responses. The inductive mode is key to knowledge generalization and better aligns with human cognition mechanisms, which is a fundamental learning mode for cognition-driven LLMs learning, thereby attracting increasing interest. Considering the importance of inductive reasoning, this survey provides a systematic exposition of inductive reasoning in LLMs, covering crucial aspects such as applications, significance, methodologies, evaluations, and analyses. Our survey is currently available at xxx.

% main paper
\section{Introduction}
\label{sec:intro}

In recent years, the rapid development of large language models (LLMs) \citep{zhao2023survey} leads to significant progress in many natural language processing (NLP) downstream tasks.
Among these, reasoning \citep{huang2022towards,DBLP:journals/corr/abs-2407-11511,zhang2025surveyreinforcementlearninglarge} is one of the comprehensive and challenging tasks for LLMs, and therefore receives considerable attention.

\begin{figure}[t!]
    \centering
    \includegraphics[width=0.5\textwidth]{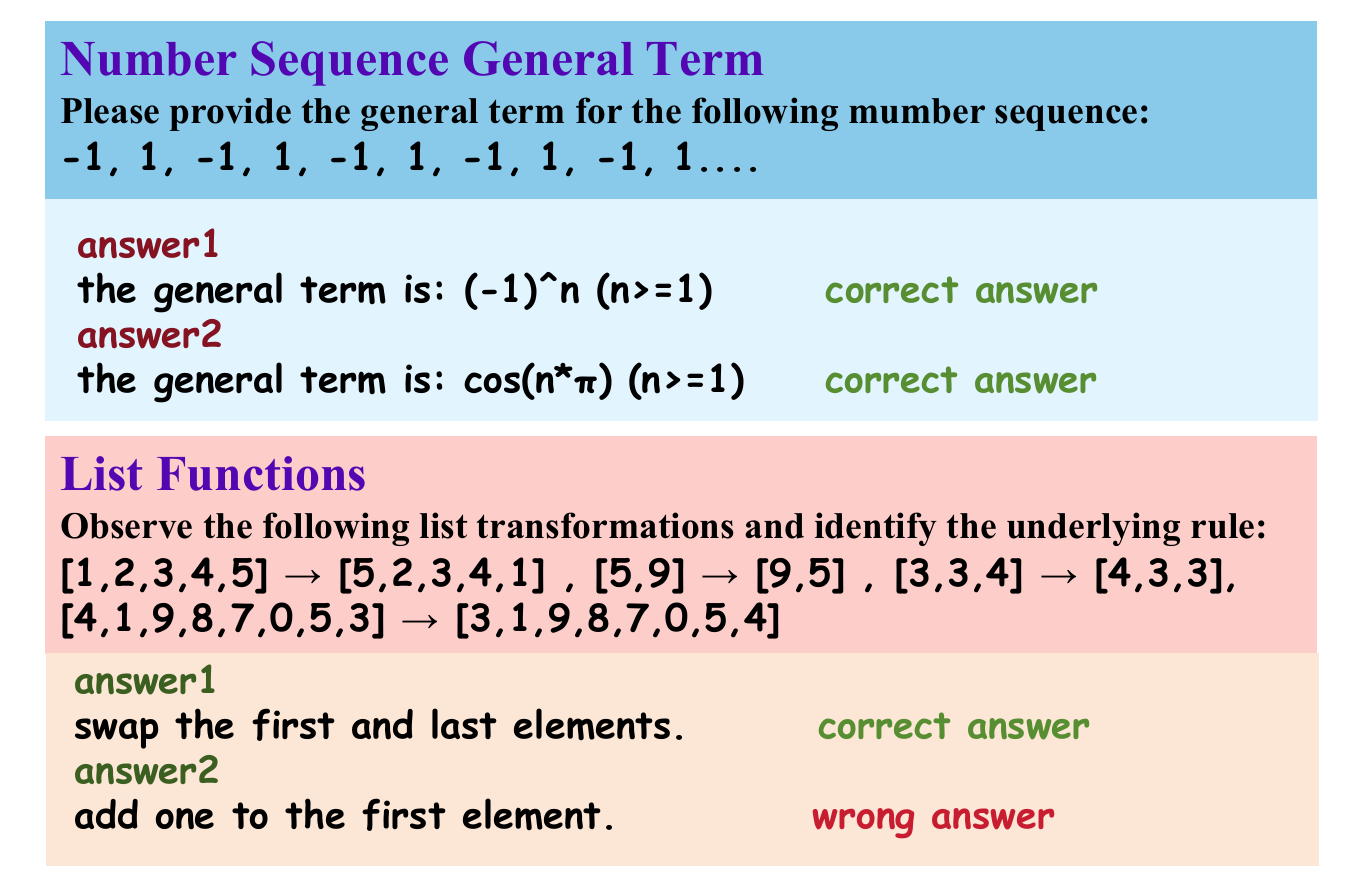} % 图片文件名
    \caption{Two examples of inductive reasoning. They generalize from specific observations or cases to derive general conclusions or rules. There may be more than one such conclusion that meets all the observations.}
    \label{fig:cases2}
\end{figure}

Within all the reasoning paradigms, inductive reasoning \citep{DBLP:conf/nips/Lu24} is one of the fundamental types.
It involves drawing general conclusions from specific observations \citep{DBLP:journals/cogsr/HanRPK24}. 
We give two examples illustrating number sequence calculation and list transformation in Figure~\ref{fig:cases2}.
The main characteristics of inductive reasoning are its particular-to-general thinking process and the non-uniqueness of its answers.
Considering how humans perceive the world, they typically make judgments by drawing analogies from past experiences to current situations, rather than always going through a strictly logical process as in deductive reasoning \citep{humanbook}.
We can assume that the inductive mode is key to knowledge generalization and better aligns with human cognition. 
It is a fundamental mode of learning and thus attracts increasing interest.

Despite the importance of inductive reasoning, previous works mostly focus on deductive reasoning \citep{li2024strategic}, represented by mathematical proof \citep{DBLP:conf/eacl/AhnVLLZY24,DBLP:journals/corr/abs-2403-00896} and program verification \citep{DBLP:conf/nips/LiuXW023,DBLP:journals/tosem/JiangDWFSLJJ24}, which is a logical reasoning that derives necessary conclusions from general rules or premises.
For more conceptual distinctions, please refer to Appendix~\ref{app: diff}.
It has already been extensively studied in recent years \citep{DBLP:journals/corr/abs-2410-08196,DBLP:conf/iclr/WangRZLLSZSZ024}.
Moreover, there is no systematic summary of inductive reasoning for LLMs.

Therefore, this paper presents the first comprehensive survey of inductive reasoning for LLMs. 
We introduce the background, including relevant concepts, applications in NLP and real-world scenarios, as well as the
significance (Section~\ref{sec: overview}) at the beginning.
The main body consists of three parts.
First, we review and prospect the methods for enhancing the inductive reasoning capabilities of LLMs (Section~\ref{sec: method}), which are categorized into three main areas: post-training enhancement, test-time exploration, and data augmentation.
We then summarize the current benchmarks of inductive reasoning and derive a unified sandbox-based evaluation approach with the observation coverage metric (Section~\ref{sec: eva}). 
Finally, we offer some theoretical analyses of inductive reasoning (Section~\ref{sec: analysis}), regarding the sources of inductive ability and practical experiences for enhancing it.
The taxonomy of this survey is in Figure~\ref{fig:taxonomy}.
In summary, the main contributions of this paper are threefold:

\begin{itemize}
\item{\textbf{First survey.} To our knowledge, we are the first to present a comprehensive survey of inductive reasoning for LLMs, thoroughly analyzing the current techniques and applications.}
\item{\textbf{New taxonomy.} We categorize methods for improving inductive reasoning into: post-training enhancement, test-time exploration, and data augmentation. We also summarize the current benchmarks and derive a unified sandbox-based evaluation approach.}
\item{\textbf{Promising direction.} We offer some analyses regarding the source of inductive ability and how simple model architectures and data help with inductive tasks, providing a solid foundation for future research (Appendix~\ref{app: future}).}
\end{itemize}

\tikzstyle{my-box}=[
    rectangle,
    draw=gray,
    rounded corners,
    text opacity=1,
    minimum height=1.5em,
    minimum width=5em,
    inner sep=2pt,
    align=center,
    fill opacity=.5,
    line width=0.8pt,
]
\tikzstyle{leaf}=[my-box, minimum height=1.5em,
    fill=pink!10, text=black, align=left,font=\normalsize,
    inner xsep=2pt,
    inner ysep=4pt,
    line width=0.8pt,
]

\definecolor{c1}{HTML}{91ADC8} % blue
\definecolor{c2}{RGB}{237,110,106} % red
\definecolor{c3}{RGB}{240,154,69} % yellow
\definecolor{c4}{RGB}{108,222,157} % green

\definecolor{c6}{RGB}{97,218,184} % cyan
\definecolor{c7}{RGB}{226,115,150} % cyan
\definecolor{c8}{RGB}{201,116,201} % cyan
\definecolor{c9}{RGB}{23,182,179} % cyan

% background
\definecolor{c15}{HTML}{FA812F} % cyan
\definecolor{c16}{HTML}{FF9013} % cyan
\definecolor{c17}{HTML}{FFC7A7} % cyan
\definecolor{c18}{HTML}{FFE8DB} % cyan
% FA812F
% FF9013
% FFC7A7
% EF7722

% methodology
\definecolor{c11}{HTML}{F39F9F}
\definecolor{c12}{HTML}{F2AEBB}

% evaluation
\definecolor{c13}{HTML}{AE75DA} 
\definecolor{c14}{HTML}{CC66DA} 
\definecolor{c5}{RGB}{205,180,243} % purple

\begin{figure*}[t]
    \centering
    \resizebox{1\textwidth}{!}{
        \begin{forest}
            forked edges,
            for tree={
                grow=east,
                reversed=true,
                anchor=base west,
                parent anchor=east,
                child anchor=west,
                base=center,
                font=\large,
                rectangle,
                % draw=hidden-draw,
                draw=gray,
                rounded corners,
                align=left,
                text centered,
                minimum width=4em,
                edge+={darkgray, line width=1pt},
                s sep=3pt,
                inner xsep=2pt,
                inner ysep=3pt,
                line width=0.8pt,
                ver/.style={rotate=90, child anchor=north, parent anchor=south, anchor=center},
                % edge path={
                %   \noexpand\path[\forestoption{edge}]
                %     (!u.parent anchor) -- ++(2mm,0)  % 先水平 5mm
                %     |- (.child anchor)\forestoption{edge label}; % 再垂直+水平进入子节点
                % },
            },
            where level=1{text width=8em,font=\normalsize,}{},
            where level=2{text width=15em,font=\normalsize,}{},
            where level=3{text width=17em,font=\normalsize,}{},
            where level=4{text width=21em,font=\small,}{},
            where level=5{text width=8em,font=\normalsize,}{},
            [
                \textbf{Inductive Reasoning for LLMs}, ver, line width=0.7mm
                [
                    \textbf{Background} (\cref{sec: overview}), fill=c16!60, draw=c16, line width=0mm
                    [   
                        \textbf{Concepts} (\cref{sec:concepts}), fill=c15!60, draw=c15, line width=0mm, edge={c15}
                        [  
                            \shortstack{\textbf{Large Language Models} (\cref{sec:llms})},  fill=c15!60, draw=c15, line width=0mm, edge={c15}
                            [
                            PLMs \citep{DBLP:journals/corr/abs-2108-05542};
                            LLMs \citep{DBLP:journals/corr/abs-2501-09171},
                            align=left, fill=c15!0, draw=c15, line width=0mm, edge={c15}
                            ]
                        ]
                        [  
                            \shortstack{\textbf{Inductive Reasoning} (\cref{sec:induc})},  fill=c15!60, draw=c15, line width=0mm, edge={c15}
                            [
                            \citet{Arthur1994InductiveRA, Heit2000PropertiesOI, Copi2004EssentialsOL, thomas2003general},
                            align=left, fill=c15!0, draw=c15, line width=0mm, edge={c15}
                            ]
                        ]
                    ]
                    [
                        \textbf{Applications} (\cref{sec:apps}), fill=c17!60, draw=c17, line width=0mm, edge={c17}
                        [
                            \shortstack{\textbf{NLP Downstream Tasks} (\cref{sec:app-nlp})},  fill=c17!60, draw=c17, line width=0mm, edge={c17}
                            [
                            Syntactic \& Semantic Parsing \citep{DBLP:conf/iclr/KimCEL20};  \hspace{2cm} \\ 
                            Information Extraction \citep{DBLP:conf/emnlp/LaiMVDN22};
                            etc.,
                            align=left, fill=c17!0, draw=c17, line width=0mm, edge={c17}
                            ]
                        ]
                        [
                            \shortstack{\textbf{Real-World Scenarios} (\cref{sec:app-real})}, fill=c17!60, draw=c17, line width=0mm, edge={c17}
                            [
                                Financial \citep{goel2025foundationtimeseriesaimodel, stempień2025hybridmodelsfinancialforecasting}; \\
                                Autonomous Driving \citep{cai2025text2scenariotextdrivenscenariogeneration,wang2025generativeaiautonomousdriving}; etc.,
                                align=left, fill=c17!0, draw=c17, line width=0mm, edge={c17}
                            ]
                        ]
                    ]
                    [
                        \textbf{Significance} (\cref{sec:significance}), fill=c18!60, draw=c18, line width=0mm, edge={c18}
                    ]
                ]
                [
                    \textbf{Methodology} (\cref{sec: method}),
                    fill=c2!60, draw=c2, line width=0mm
                    [
                        \textbf{Post-training Enhancement} (\cref{sec: post-train}), align=center, fill=c7!60, draw=c7, line width=0mm, edge={c7}
                        [
                            \shortstack{\textbf{Synthetic Data} (\cref{sec:syndata})}, align=center, fill=c7!60, draw=c7, line width=0mm, edge={c7}
                            [
                                LingR \citep{jiang-etal-2024-linguistic};
                                ItD \citep{sun2024itdlargelanguagemodels};   \hspace{2cm} \\
                                CodeSeq \citep{chen2025codedriveninductivesynthesisenhancing}; etc.,
                                align=left, fill=c7!0, draw=c7, line width=0mm, edge={c7}
                            ]
                        ]
                        [
                            \shortstack{\textbf{IRL-style Optimization}  (\cref{sec:irl})}, align=center, fill=c7!60, draw=c7, line width=0mm, edge={c7}
                            [
                                IRL \citep{arora2021survey};
                                RLHF \citep{jin2025rewardfunctionrlbest};   \hspace{2cm} \\
                                Prompt-OIRL \citep{sun2024querydependentpromptevaluationoptimization};
                                etc.,
                                align=left, fill=c7!0, draw=c7, line width=0mm, edge={c7}
                            ]
                        ]
                    ]
                    % TODO:
                    [  
                        \textbf{Test-time Exploration} (\cref{sec: hypo}), fill=c11!60, draw=c11, line width=0mm, edge={c11}
                        [
                            \shortstack{\textbf{Hypothesis Selection}  (\cref{sec: hyposel})}, fill=c11!60, draw=c11, line width=0mm, edge={c11}
                            [
                            moc \citep{DBLP:conf/naacl/LeeKLYKJ25};
                            epic \citep{DBLP:conf/acl/ParfenovaP25};
                            etc.,
                                align=left, fill=c11!0, draw=c11, line width=0mm, edge={c11}
                            ]
                        ]
                        [
                            \shortstack{\textbf{Hypothesis Iteration}  (\cref{sec: hypoite})}, fill=c11!60, draw=c11, line width=0mm, edge={c11}
                            [
                            ARISE
                            \citep{DBLP:conf/naacl/MSMKR25};
                            SSR \citep{DBLP:conf/acl/Li0ZS25}; \hspace{2cm} \\
                            IDEA \citep{he2025ideaenhancingrulelearning};
                            etc.,
                                align=left, fill=c11!0, draw=c11, line width=0mm, edge={c11}
                            ]
                        ]
                        [
                            \shortstack{\textbf{Hypothesis Evolution}  (\cref{sec: hypoevo})}, fill=c11!60, draw=c11, line width=0mm, edge={c11}
                            [
                             HRI \citep{DBLP:conf/icml/GlanoisJFWZ0LH22};
                             IncSchema \citep{li-etal-2023-open};   \hspace{2cm} \\
                             PRIMO \citep{liu2024primoprogressiveinductionmultihop};
                            etc.,
                                align=left, fill=c11!0, draw=c11, line width=0mm, edge={c11}
                            ]
                        ]
                    ]
                    [
                        \textbf{Data Augmentation} (\cref{sec: info}), fill=c12!60, draw=c12, line width=0mm, edge={c12}
                        [
                            \textbf{Human Intervention} (\cref{sec:human}), fill=c12!60, draw=c12, line width=0mm, edge={c12}
                            [
                            SS-VQ-VAE \citep{huang-ji-2020-semi};
                            Sc.WSI \citep{eyal-etal-2022-large};
                              \hspace{2cm} \\
                            HITL-SI \citep{zhang-etal-2023-human};
                            etc.,
                                align=left, fill=c12!0, draw=c12, line width=0mm, edge={c12}
                            ]
                        ]
                        [
                            \shortstack{\textbf{External Knowledge Retrieval} (\cref{sec:external})}, fill=c12!60, draw=c12, line width=0mm, edge={c12}
                            [
                            LLEGO \citep{DBLP:conf/iclr/LiuHS25};
                            iCoT \cite{DBLP:conf/emnlp/ChenWHC24};
                            \hspace{2cm}
                            \\
                            CommExpl \citep{DBLP:conf/acl/RyuSFXPH22}
                            etc.,
                                align=left, fill=c12!0, draw=c12, line width=0mm, edge={c12}
                            ]
                        ]
                        [
                            \shortstack{\textbf{Structured Signals} (\cref{sec:structure})}, fill=c12!60, draw=c12, line width=0mm, edge={c12}
                            [
                            QARR \citep{DBLP:conf/acl/XieZ0ZH23};
                            REST \citep{DBLP:journals/corr/abs-2408-07088}; \hspace{2cm} \\
                            GI-LUG \citep{DBLP:conf/emnlp/KaiHHL24}
                            etc.,
                                align=left, fill=c12!0, draw=c12, line width=0mm, edge={c12}
                            ]
                        ]
                    ]
                ] 
                [
                    \textbf{Evaluation} (\cref{sec: eva}), fill=c14!60, draw=c14, line width=0mm
                    [
                        \textbf{Benchmarks} (\cref{sec: benchmark}), fill=c13!60, draw=c13, line width=0mm, edge={c13}
                        [
                        ARC \citep{DBLP:journals/corr/abs-1911-01547};
                        List Functions \citep{rule2020child};
                        ILP \citep{evans2018learningexplanatoryrulesnoisy};
                        DEER \citep{DBLP:conf/eacl/YangDDCCLGW24}; etc. \hspace{1cm}, 
                        align=left, fill=c13!0, draw=c13, line width=0mm, edge={c13}
                        ,text width=39.5em, font=\small,
                        ]
                    ]     
                    [
                        \textbf{Evaluation Approaches} (\cref{sec:evalm}), fill=c5!60, draw=c5, line width=0mm, edge={c5}
                        [
                            \shortstack{\textbf{Existing Evaluation Strategy} (\cref{sec: traeva})}, fill=c5!60, draw=c5, line width=0mm, edge={c5}
                            [
                            ACRE \citep{DBLP:conf/cvpr/0017JEZZ21};
                            SCAN \citep{lake2018generalizationsystematicitycompositionalskills}; \hspace{2cm} \\
                            SyGuS \citep{DBLP:conf/iclr/OdenaSBSSD21};
                            etc.,
                                align=left, fill=c5!0, draw=c5, line width=0mm, edge={c5}
                            ]
                        ]
                        [
                            \shortstack{\textbf{Sandbox-based Evaluation} (\cref{sec: sandbox})}, fill=c5!60, draw=c5, line width=0mm, edge={c5}
                            [
                            Code \citep{DBLP:journals/corr/abs-2501-01277};
                            Tool \citep{DBLP:journals/fcsc/QuDWCWYXW25}; \hspace{2cm} \\ 
                            LLM-as-a-Judge \citep{DBLP:journals/corr/abs-2411-15594};
                            etc.,
                                align=left, fill=c5!0, draw=c5, line width=0mm, edge={c5}
                            ]
                        ]
                    ]  
                ]% level1
                [
                    \textbf{Analysis} (\cref{sec: analysis}), fill=c1!60, draw=c1, line width=0mm
                ]
            ]
        \end{forest}
    }
    % \vspace{-0mm}
    \caption{Taxonomy of the survey about the inductive reasoning for LLMs.}
    \label{fig:taxonomy}
    % \vspace{0mm}
\end{figure*}
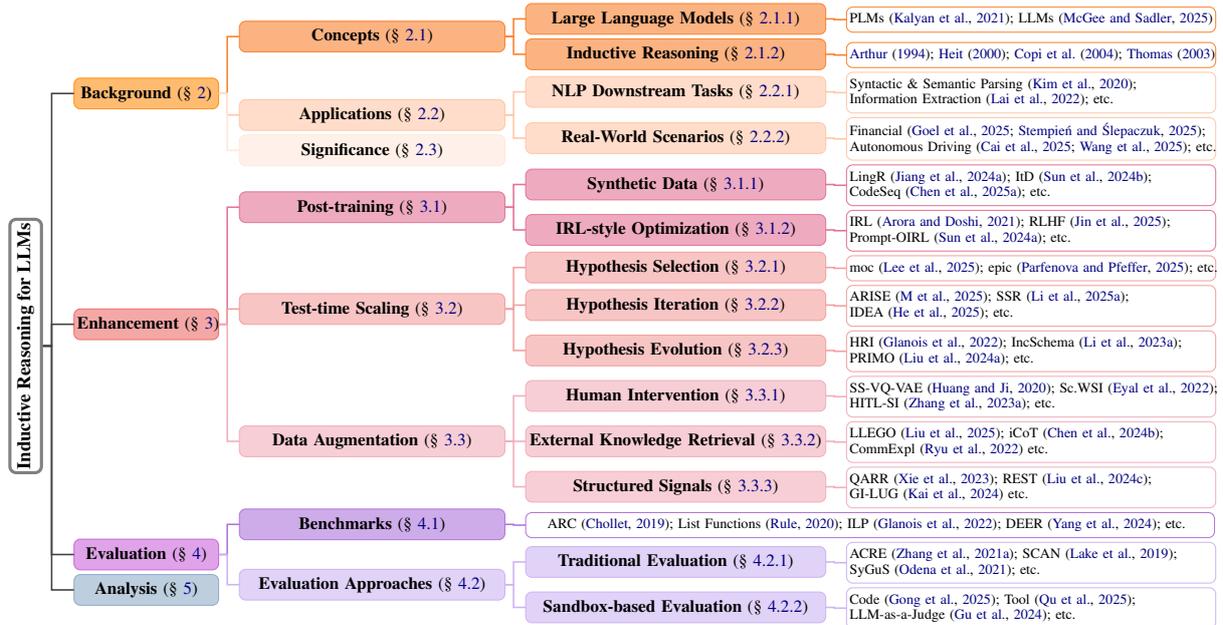

\section{Background}
\label{sec: overview}
In this section, we will introduce the relevant concepts of inductive reasoning, some of its application scenarios, and the significance of studying it.

\subsection{Concepts}
\label{sec:concepts}
\subsubsection{Large Language Models}
\label{sec:llms}
Since the transformer architecture \citep{DBLP:conf/nips/VaswaniSPUJGKP17} has become mainstream for language models, the field of NLP has experienced rapid development \citep{DBLP:journals/corr/abs-2110-08455}.
During $2017$ to $2022$, pretrained language models (PLMs) \citep{DBLP:journals/corr/abs-2108-05542}, which undergo two stages---pretraining and finetuning---such as the BERT \citep{devlin2019bertpretrainingdeepbidirectional} and T5 \citep{raffel2023exploringlimitstransferlearning} series, once dominated the entire field.
From $2022$, with the advent of ChatGPT-$3.5$ \citep{DBLP:journals/corr/abs-2303-10420}, the era of LLMs officially begins.
LLMs, with their massive number of parameters and unique training methods \citep{zhao2025surveylargelanguagemodels}, significantly improve the performance of NLP tasks \citep{DBLP:journals/corr/abs-2311-07194,DBLP:journals/corr/abs-2312-17617,DBLP:journals/corr/abs-2407-11511} and have a profound impact on various aspects of daily life \citep{DBLP:journals/corr/abs-2311-13160,DBLP:journals/corr/abs-2311-05112,DBLP:journals/corr/abs-2308-06013}.
Some well-known LLMs include the GPT series\footnote{https://chatgpt.com/overview}, the Gemini series\footnote{https://cloud.google.com/vertex-ai/generative-ai/docs}, the Claude series\footnote{https://claude.ai}, and so on.

\subsubsection{Inductive Reasoning}
\label{sec:induc}
Inductive reasoning represents \textit{\textbf{making an induction from specific instances or observations to derive general rules and conclusions}} \citep{Arthur1994InductiveRA, Heit2000PropertiesOI}.
From another perspective, it denotes one reasoning approach where the conclusion is not guaranteed with certainty, but instead supported only to a certain degree of probability \cite{Copi2004EssentialsOL}.
In other words, inductive reasoning may have more than one valid hypothesis that can account for all the instances or observations, making its answer open \citep{thomas2003general}.
To sum up abstractly, inductive reasoning is a thought process that proceeds from the particular to the general.

\subsection{Applications of Inductive Reasoning}
\label{sec:apps}
The core idea of inductive reasoning is inductive bias.
It is a set of assumptions or prior conditions that a model or an individual relies on when encountering unseen items \citep{Caruana1993MultitaskLA,Baxter2000AMO}.
It is hard to find a `universal’ bias in deep learning. 
Choosing an appropriate inductive bias for a specific task is key to achieving success \citep{Provost1995InductivePT}.
We will discuss its applications in NLP tasks and real-world scenarios.

\subsubsection{NLP Downstream Tasks}
\label{sec:app-nlp}
Inductive reasoning is widely applied to improve the performance of NLP downstream tasks (Appendix~\ref{app: nlp}). 
Some common practices include training models to learn inductive bias \citep{DBLP:journals/ijcv/YangZCX25}, constructing chains of thought (CoT) \citep{chen2025reasoningerasurveylong} or summarizing rules to enhance interpretability \citep{DBLP:journals/inffus/XuY25}, and leveraging intra-parameters implicit knowledge \citep{cheng2024understandinginterplayparametriccontextual} for induction, and others. 
Such approaches benefit a wide range of downstream NLP tasks:
syntactic and semantic parsing \citep{DBLP:conf/iclr/KimCEL20,DBLP:conf/acl/YamadaST21,DBLP:conf/emnlp/LindemannKT24,tsujimoto-etal-2025-semantic},
information extraction \citep{DBLP:conf/emnlp/LaiMVDN22,DBLP:journals/corr/abs-2408-07088,silva-etal-2025-inductive,DBLP:conf/eacl/XuZC24}, dialogue systems \cite{DBLP:conf/acl/FengR0022,DBLP:conf/acl/XieMZMYWM24,DBLP:conf/emnlp/OuWLZZG24}, 
question answering \cite{DBLP:conf/coling/Gu022,DBLP:conf/emnlp/KimHMW23,DBLP:conf/acl/ChenLZHZ24}, and multimodal tasks \cite{DBLP:conf/emnlp/AmosyVSBRC24,DBLP:journals/corr/abs-2503-07565,DBLP:journals/corr/abs-2501-16524}.

\subsubsection{Real-World Scenarios}
\label{sec:app-real}
Inductive reasoning has a broad impact on real-world scenarios.
We list three of them.
(1) Financial forecasting: Inductive models are essential for predicting future financial outcomes by learning complex, non-linear patterns from vast amounts of historical time-series data \citep{faheem2021ai,goel2025foundationtimeseriesaimodel, stempień2025hybridmodelsfinancialforecasting}.
(2) Autonomous driving: 
Inductive reasoning enables autonomous driving systems to generalize from historical data and past experiences to recognize rules in traffic conditions to make decisions\citep{cai2025text2scenariotextdrivenscenariogeneration,wang2025generativeaiautonomousdriving}.
(3) Conversational healthcare and diagnostic dialogue:
Inductive reasoning empowers artificial intelligence systems to mimic a clinician’s process of taking patient history and formulating a diagnosis \citep{zhu2025cpenvevaluatinglargelanguage,zhu2026medmcpcalcbenchmarkingllmsrealistic} by generalizing from symptom patterns \citep{tu2024conversationaldiagnosticai,dhudum2024revolutionizing,zhang2025graphneuralnetworksmodern}.
More broader real-world applications are extended in Appendix~\ref{app: future}.

\subsection{Significance of Inductive Reasoning}
\label{sec:significance}
The broad applications of inductive reasoning in both AI and real‑world scenarios demonstrate its universality and fundamental contribution to knowledge discovery and generalization \citep{DBLP:journals/tkde/CarterH98,DBLP:conf/www/BaiFBTLHT24,DBLP:conf/kdd/2025-1}: 
(1) Deriving general conclusions from specific cases, allowing it to cover and generalize to a wider range of applications, which aligns with the human learning process. 
(2) Adaptive adjustments help in uncertain and complex scenarios, where inductive reasoning may yield multiple plausible outcomes rather than a single unique solution.

\section{Methodology}
\label{sec: method}

In this section, we will introduce three major approaches to enhance the inductive capabilities of LLMs: post-training enhancement(Section~\ref{sec: post-train}), test-time exploration (Section~\ref{sec: hypo}), and data augmentation (Section~\ref{sec: info}).
It is worth noting that we not only summarize existing methods but also prospect a forward-looking review of potential future inductive approaches.
For convenience, we treat the inputs of inductive reasoning as observations and refer to these outputs as rules.
Boundaries and comparisons of the three methods can be found in the Appendix~\ref{app: med}.

\subsection{Post-training Enhancement}
\label{sec: post-train}

\begin{figure*}[t]
  \centering
  \subfloat[synthetic data]{\includegraphics[width=0.32\linewidth]{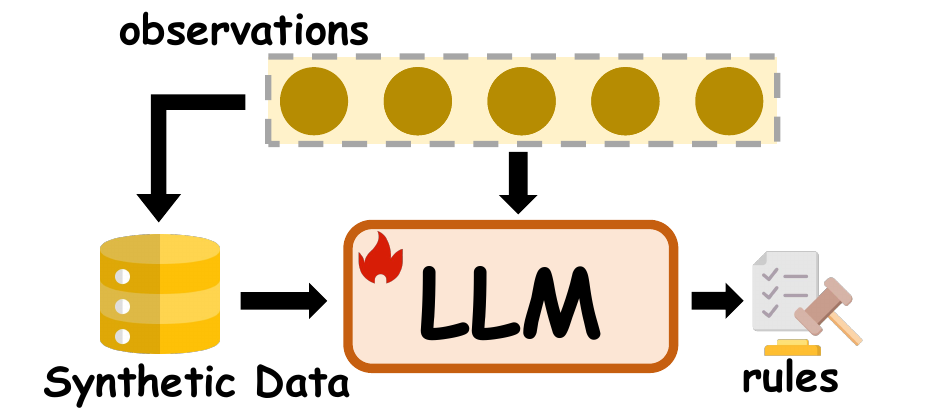}}
  \quad
  \subfloat[IRL-style optimization]{\includegraphics[width=0.32\linewidth]{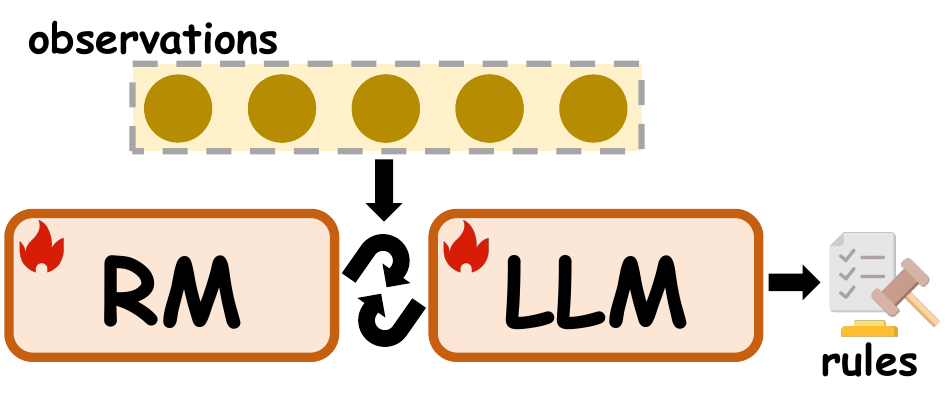}}
  \caption{The demonstration of the post-training enhancement method for inductive reasoning.}
  \label{fig: post-training}
\end{figure*}

Post-training enhancement refers to improving the inductive reasoning ability of LLMs during the post-training stage \citep{DBLP:conf/acl/LaiLGCQXYZDHLHD25,DBLP:journals/corr/abs-2505-02686}, using algorithms such as supervised finetuning (SFT) and reinforcement learning (RL).
This category of methods primarily focuses on constructing synthetic data \citep{DBLP:journals/corr/abs-2406-15126,DBLP:conf/kdd/JiangLC25} (Section~\ref{sec:syndata}) and developing new algorithms (Section~\ref{sec:irl}).
We illustrate them in Figure~\ref{fig: post-training}.

\subsubsection{Synthetic Data}
\label{sec:syndata}
Synthetic data means artificially generated data that mimics the properties and patterns of real-world data \citep{bauer2024comprehensive}.
Data plays a decisive role in LLM training.
To address certain inherent limitations of natural data, such as being difficult to obtain or organize \citep{DBLP:journals/access/NadasDT25}, researchers often construct data manually to compensate for these shortcomings.
LingR \citep{jiang-etal-2024-linguistic} builds a `linguistic rule instruction set' for various LLMs, enabling them to learn step-by-step reasoning based on linguistic rules such as causality.
ItD \citep{sun2024itdlargelanguagemodels} leverages the deductive abilities of LLMs to generate data and optimize inductive ability. 
The model’s capacity to learn general rules from a small number of samples is significantly enhanced.
CodeSeq \citep{chen2025codedriveninductivesynthesisenhancing} constructs SFT and RL training sets to ask LLMs to facilitate reasoning over number sequence general term formulas, thereby improving their inductive abilities.
Other approaches \citep{wu2022limelearninginductivebias,aksu2023cesarautomaticinductioncompositional,darm2023knowledgebasequestionanswering,mosolova-etal-2025-llm,DBLP:conf/iclr/LiCJ00025} establish similar induction-related training datasets for the models to learn from.

\subsubsection{IRL-style Optimization}
\label{sec:irl}
Reward models (RMs) \citep{zhong2025comprehensive} are typically utilized to provide supervision signals for the RL process. 
However, for inductive reasoning, due to the non-uniqueness of answers and the uncertainty in the reasoning process, traditional RMs struggle to provide effective supervision. Therefore, Inverse RL \citep{arora2021survey} (IRL), which needs to induce the latent reward functions, may serve as an alternative approach \citep{sun2025inversereinforcementlearningmeets}.
The Reinforcement Learning from Human Feedback (RLHF) \citep{kaufmann2024survey,swamy2024minimaximalist} can be seen as a process of IRL to some extent, as it infers human preferences and the underlying reward function from human feedback essentially.
Therefore, designing an appropriate reward model in RLHF can enhance the inductive reasoning ability of LLMs \citep{jin2025rewardfunctionrlbest}.
We can also employ Prompt-OIRL \citep{sun2024querydependentpromptevaluationoptimization} for reference, which proposes an IRL-based method that reuses historical prompting trial-and-error experience to train a reward model to improve the model’s inductive exploration ability.
Although works combining IRL and reasoning are still scarce, the approach of IRL---fitting the posterior distribution of the reward model from human or data signals \citep{cai2024approximatedvariationalbayesianinverse,krishna2024solvinginversealignmentproblem} ---has strong extensibility and can thus be regarded as one of the important methods for the facilitation of inductive reasoning.

\subsection{Test-time Exploration}
\label{sec: hypo}

\begin{figure*}[t]
  \centering
  \subfloat[hypothesis selection]{\includegraphics[width=0.31\linewidth]{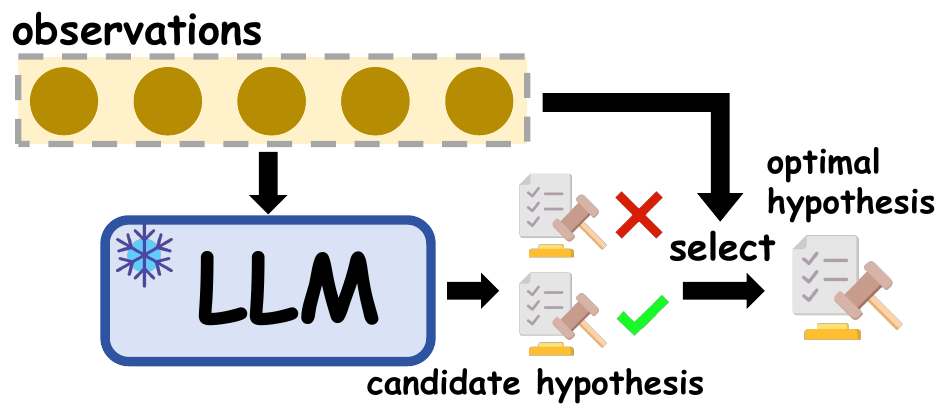}}
  \quad
  \subfloat[hypothesis iteration]{\includegraphics[width=0.31\linewidth]{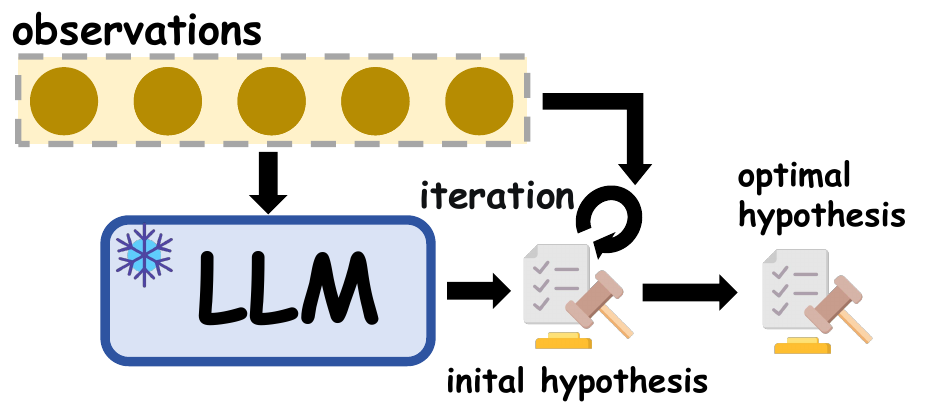}}
  \quad
  \subfloat[hypothesis evolution]{\includegraphics[width=0.31\linewidth]{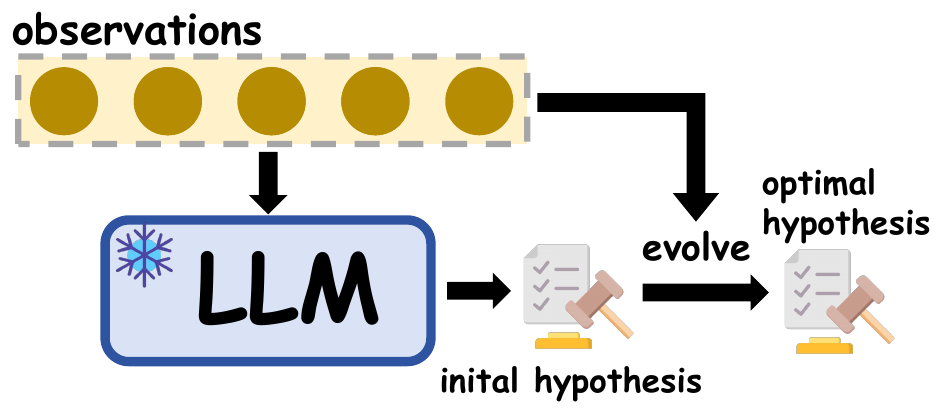}}
  \caption{The demonstration of the test-time exploration method for inductive reasoning.}
  \label{fig: hypothesis}
\end{figure*}

The goal of inductive reasoning is to derive general rules from observations, which inevitably involves forming hypotheses during the reasoning process. 
The above post-training enhancement requires training LLMs, whereas the test-time exploration in this section is a hypothesis-based method that only works during the inference stage \citep{DBLP:journals/corr/abs-2503-24235}.
Unlike end-to-end models \citep{DBLP:conf/ijcai/KotaryFHW21}, the test-time exploration method prompts the frozen LLMs to form an inductive reasoning pipeline \citep{he2025reasoninglearningsurveyhypothesis}.
We have LLMs generate candidate hypotheses for inductive problems, which can then undergo selection (Section~\ref{sec: hyposel}), iteration (Section~\ref{sec: hypoite}), or evolution (Section~\ref{sec: hypoevo}) operations to reach the optimal one. 
Detailed processes are shown in Figure~\ref{fig: hypothesis}.

\subsubsection{Hypothesis Selection}
\label{sec: hyposel}
Hypothesis selection refers to choosing, from the candidate hypotheses generated by LLMs, those that can cover the observations \citep{pazzani1990feature,bun2019private}.
Hypothesis Search \citep{DBLP:conf/iclr/WangZPPHG24} let the LLMs generate multiple abstract hypotheses in natural language. Then, narrow down the hypothesis set through either the LLMs or minimal human filtering.
The motivation of Mixture of Concepts (MoC) \citep{DBLP:conf/naacl/LeeKLYKJ25} lies in the fact that hypotheses for inductive reasoning often produce semantic redundancy. 
Therefore, the proposed method simulates human inductive reasoning by first figuring out a list of semantically non-redundant concepts and then generating corresponding hypotheses based on each concept.
\citet{DBLP:conf/acl/ParfenovaP25} proposes Ensemble Pipeline for Inductive Coding (EPIC) to address the issue of inconsistency in inductive encodings by using small LLMs to generate candidate encodings, filtering them through a moderator mechanism and similarity checks, and finally evaluating them with composite metrics.

\subsubsection{Hypothesis Iteration}
\label{sec: hypoite}
Hypothesis iteration means iterating over candidate hypotheses until they satisfy all the observations \citep{yom2015methodology}.
\citet{qiu2024phenomenalpuzzlingtestinginductive} proposes a three-step iterative hypothesis refinement method that simulates the human inductive reasoning process: generate multiple hypotheses from a few examples; evaluate how many known instances each hypothesis can cover; and have the LLMs further revise the selected hypotheses based on feedback, iterating for several rounds until convergence.
SSR \citep{DBLP:conf/acl/Li0ZS25} iteratively optimizes by generating diverse candidate rules and refining them based on execution feedback.
ARISE \citep{DBLP:conf/naacl/MSMKR25} also iterates over the inductive rules before using them to train the model.
IDEA framework \citep{he2025ideaenhancingrulelearning} resolves the shortcomings of LLMs in interactive rule learning by simulating the human cycle of hypothesis revision, thereby enhancing the model’s dynamic learning capability.

\subsubsection{Hypothesis Evolution}
\label{sec: hypoevo}
Unlike the iterative process, hypothesis evolution expands, diversifies, or evolves the hypothesis space by generating, filtering, and combining multiple hypotheses, forming hypotheses that better capture complex patterns \citep{DBLP:journals/kybernetes/Galkin11,DBLP:conf/aaai/GilGRMABSM17,DBLP:journals/entropy/Juretic25}.
LLMs leverage contextual and label information, along with prompts, to progressively guide the model in dynamically generating patterns during reasoning, without relying on predefined rules \citep{DBLP:conf/eacl/DrorWR23}.
IncSchema \citep{li-etal-2023-open} gradually induces general patterns by querying the LLMs in stages---first listing core events, then expanding details, and finally verifying relationships.
HRI \citep{DBLP:conf/icml/GlanoisJFWZ0LH22} generates inductive meta-rules and matches them with samples, thereby evolving into first-order logic rules.
PRIMO \citep{liu2024primoprogressiveinductionmultihop} introduces a progressive multi-stage open rule induction method for deriving multi-hop rules, thereby capturing more complex reasoning chains.

\subsection{Data Augmentation}
\label{sec: info}
\begin{figure*}[t]
  \centering
  \subfloat[human intervention]{\includegraphics[width=0.31\linewidth]{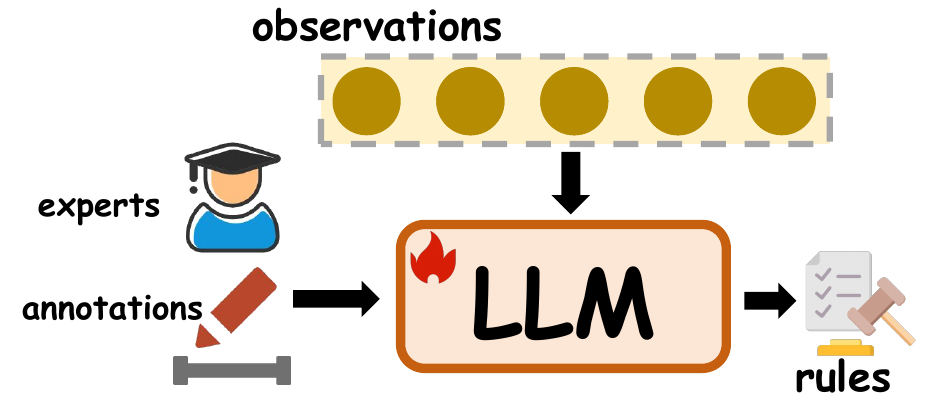}}
  \quad
  \subfloat[external knowledge]{\includegraphics[width=0.31\linewidth]{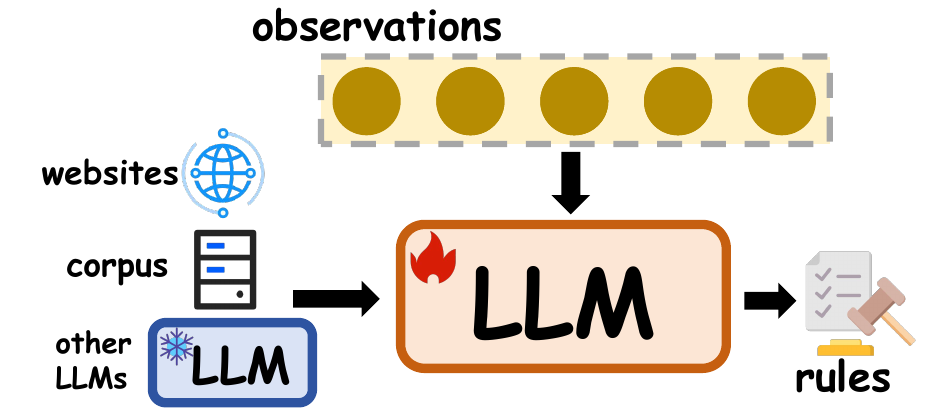}}
  \quad
  \subfloat[structured signals]{\includegraphics[width=0.31\linewidth]{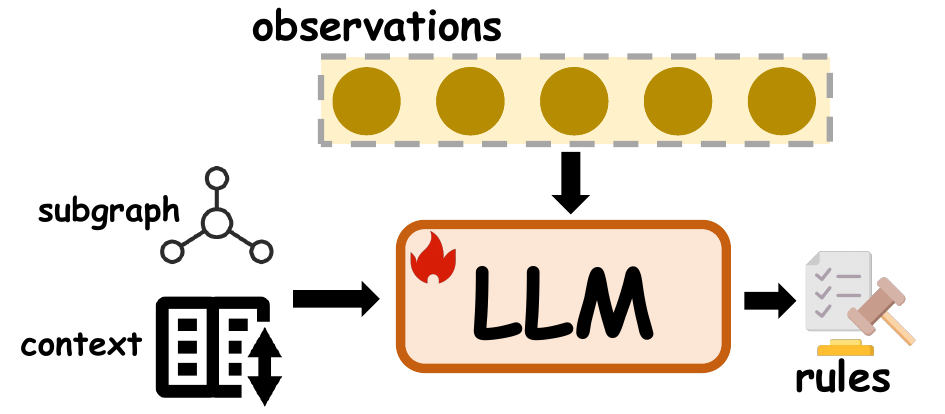}}
  \caption{The demonstration of the data augmentation method for inductive reasoning.}
  \label{fig: information augmentation}
\end{figure*}

Data augmentation \citep{DBLP:conf/sigir/ZhangF022} for LLMs signifies enriching the model’s input with additional knowledge or structured signals, such as external facts and retrieved documents, to enhance reasoning and improve output quality.
We divide it into three subcategories: human intervention (Section~\ref{sec:human}), external knowledge (Section~\ref{sec:external}), and structured signals (Section~\ref{sec:structure}).
Please refer to Figure~\ref{fig: information augmentation} about them.

\subsubsection{Human Intervention}
\label{sec:human}
Human intervention incorporates expert knowledge or human-annotated information during inductive reasoning.
SS-VQ-VAE \citep{huang-ji-2020-semi} relies on a small amount of human-annotated information to discover new patterns.
\citet{eyal-etal-2022-large} generates substitute words, then annotates the corpus and trains static embeddings to enhance the model’s inductive priors.
\citet{zhang-etal-2023-human} utilizes GPT-$3$ to generate candidate patterns and enhances their quality through human intervention, addressing the issues of domain transfer and semantic consistency in purely automated approaches.
Some other studies \citep{DBLP:conf/eacl/EdwardsJ23,DBLP:conf/naacl/VerhoevenMBYS24} emphasize inductive capabilities in low-annotation scenarios, which indirectly reflects the importance of expert knowledge.

\subsubsection{External Knowledge}
\label{sec:external}
In this paper, we define external knowledge \citep{DBLP:conf/acl/CaoZ000CP20} to include web or document information, knowledge from other corpora, knowledge stored in LLM parameters \citep{DBLP:journals/corr/abs-2204-06031}, and so on.
LLEGO \citep{DBLP:conf/iclr/LiuHS25} incorporates the semantic prior knowledge embedded in large LLMs into genetic programming operations to enhance generalization ability.
The parameter knowledge of LLMs \citep{DBLP:conf/emnlp/ZhangZRSHWLC23} and multimodal large models \citep{DBLP:conf/naacl/ZhangSJXYL21,li2024reevaluatingneedmultimodalsignals} can also serve as an important supplementary information source for inductive tasks.
For example, some powerful LLMs are directly prompted to produce the inductive Chain-of-Thought \citep{DBLP:conf/emnlp/ChenWHC24}, inductive steps \citep{DBLP:conf/acl/QianWLLY23}, and inductive rules \citep{DBLP:conf/acl/RamjiR25} for the current task, providing additional assistance.
Other types of knowledge, such as bilingual corpora \citep{DBLP:conf/acl/ShiZW20,DBLP:conf/naacl/KohliFDMMK24}, social media content \citep{DBLP:conf/aclnut/RadhakrishnanKC20}, and commonsense knowledge \citep{DBLP:conf/acl/RyuSFXPH22}, can be used for inductive tasks in the same way.

\subsubsection{Structured Signals}
\label{sec:structure}
Structured information represents subgraphs or contextual information of LLMs ( neighboring hidden states or embeddings), which provide local implicit signals and help LLMs to learn relevant inductive biases \citep{DBLP:conf/acl/ImmerHFC22}.
\citet{DBLP:conf/emnlp/LiKV23} optimizes the model’s output by retrieving nearest-neighbor embeddings as contextual examples.
QARR \citep{DBLP:conf/acl/XieZ0ZH23} extracts an open subgraph for the query entity to inductively infer new entities.
REST \citep{DBLP:journals/corr/abs-2408-07088} deploys rule-induced subgraphs to capture local semantic patterns, thereby enhancing the model’s generalization ability in inductive scenarios.
GI-LUG \citep{DBLP:conf/emnlp/KaiHHL24} uses a syntactic parser to generate syntax masks that guide the attention mechanism, and combine the byte pair encoding (BPE) embeddings with a hybrid loss function to optimize the induction process.
Although this type of method is widely used in the PLM era, due to the same underlying principle, it can also play an important role for LLMs.

\section{Evaluation}
\label{sec: eva}
In this section, we will introduce current benchmarks for LLM inductive reasoning, some evaluation approaches, and the corresponding metrics.

\subsection{Benchmarks}
\label{sec: benchmark}

\begin{table*}[t]
  \centering
  \small
  \renewcommand{\arraystretch}{1.2} % 调整行间距为 1.1 倍
  \caption{Some benchmarks for evaluating the inductive reasoning abilities of LLMs. We provide the atomic objects, names with their references, the input formations, the targets to be induced, and the number of test samples (approximate values). `.' represents that it is the abbreviation of the benchmark name, while `*' indicates that the data are presented in the form of analogical reasoning. Further details about these benchmarks are in Appendix~\ref{app: bench}.}
  \label{tab: bench}
  \begin{tabular}{llp{3.5cm}p{3.5cm}p{1.5cm}}
    \toprule
     \textbf{Object} & \textbf{Benchmark Name} &\textbf{Obervation Input} &  \textbf{Induction Target} &  \textbf{\# Samples}\\
    \midrule
    symbol &  ILP \citep{evans2018learningexplanatoryrulesnoisy} & pos. and neg. samples & a one-order logic rule  & $1,500$\\
    entity &  SCAN
    \citep{lake2018generalizationsystematicitycompositionalskills} & a state of entities   &  an action of the state & $4,000$\\
    grid &  ARC*
    \citep{DBLP:journals/corr/abs-1911-01547} & pairs of grids   & a grid conversion rule & $400$\\
    list &  List Func.*
    \citep{rule2020child} & pairs of number lists &  a list operation rule & $1,200$\\
    code & PROGES \citep{DBLP:conf/icml/AletLKNSLKT21} & IO input/output & a program & $270,000$\\
    string & SyGuS \citep{DBLP:conf/iclr/OdenaSBSSD21} & a pair of strings & a string-mapping program & $8,00$\\
    entity &  ACRE
    \citep{DBLP:conf/cvpr/0017JEZZ21} & functions of entities &  a `Blickets' entity & $6,000$\\
    text  & Instruc. \citep{honovich2022instructioninductionexamplesnatural} & two NL sentences & an instruction & $2,400$\\
    number &  Arith.*
    \citep{DBLP:conf/naacl/WuQRA0WKAK24} & two numbers & the sum in certain base & $1,000$\\
    symbol &  Le/Ho.
    \citep{DBLP:conf/emnlp/LiuLCS24} & pairs of triplets & an entailment rule  & $2,000$\\
    structure & NutFrame \citep{DBLP:conf/coling/Guo000024} & some frame information & conceptual structures &$30,000$ \\
    fact & DEER \citep{DBLP:conf/eacl/YangDDCCLGW24} & a pair of facts & a rule covers the facts &$200$\\
    puzzle & RULEARN \citep{he2025ideaenhancingrulelearning} & some puzzle scenarios & a puzzle rule &$300$\\
    word &  Crypto.*
    \citep{DBLP:conf/acl/Li0ZS25} & pairs of english words &  an encrypted rule &$300$\\
    symbol &  GeoILP
    \citep{DBLP:conf/iclr/ChenZZ25} & pos. and neg. samples &  a logic rule &$250,000$\\
    string &  In.Bench
    \citep{DBLP:conf/acl/HuaWSPJW25} & a pair of strings & a string-mapping rule &$1,000$\\
    number & CodeSeq \citep{chen2025codedriveninductivesynthesisenhancing} & a number sequence & the general term & $200$ \\
    \bottomrule
  \end{tabular}
\end{table*}

To evaluate the inductive reasoning capabilities of LLMs, the research community constructs a diverse set of benchmarks, as shown in Table~\ref{tab: bench}, that comprehensively assess the models' ability to generalize universal rules from concrete observations.
It is noteworthy that the input formats of some tasks appear as paired samples or few-shot examples, often framed as analogy reasoning. 
As we claim in Appendix~\ref{app: diff}, since analogical reasoning is a special form of inductive reasoning, we also regard benchmarks in this analogical form as benchmarks for inductive reasoning.
The core task of these inductive benchmarks requires models to observe a small number of input examples (Observation Input), infer underlying patterns, and output the final rules (Induction Target).

As shown in Table~\ref{tab: bench}, the data objects covered by these benchmarks span a wide range---from basic structures such as numbers, strings, and lists, to more complex forms like grids, logical formulas, and even natural language text.

Among them, benchmarks such as ARC \citep{DBLP:journals/corr/abs-1911-01547}, List Functions \citep{rule2020child}, and SyGuS \citep{DBLP:conf/iclr/OdenaSBSSD21} focus on algorithmic or rule learning, requiring models to generate programs or operational rules that explain data transformations. 
What's more, tasks like ILP \citep{evans2018learningexplanatoryrulesnoisy}, GeoILP  \citep{DBLP:conf/iclr/ChenZZ25}, and ACRE \citep{DBLP:conf/cvpr/0017JEZZ21} place greater emphasis on the induction of logical concepts and symbolic rules.
Particularly, Codeseq \citep{chen2025codedriveninductivesynthesisenhancing} involves the computation of number sequence general terms, which represents a more advanced and complex form of inductive reasoning.

Overall, these benchmarks test the models’ pattern recognition ability and impose rigorous challenges on their higher-order cognitive skills. 
They not only examine how effectively LLMs can generalize from limited observations to underlying rules, but also serve as a foundation for driving further progress in enhancing such abilities.

\subsection{Evaluation Approaches}
\label{sec:evalm}
In this section, we introduce the evaluation approaches for the inductive reasoning benchmarks mentioned above. 
We first present the existing evaluation strategies used in the benchmark papers (Section~\ref{sec: traeva}). 
Then, we derive a sandbox-based evaluation approach with a fine-grained observation coverage metric built upon it (Section~\ref{sec: sandbox}).

\subsubsection{Existing Evaluation Strategy}
\label{sec: traeva}

Most of the benchmarks in Section~\ref{sec: benchmark} and current works directly evaluate the consistency between answers generated by LLMs and the ground truth.
Therefore, general metrics are employed, such as ACC, exact match, success rate, and so on.
For example, ACRE \citep{DBLP:conf/cvpr/0017JEZZ21} selects the most plausible “Blicket” from multiple options to evaluate the accuracy of its selection.
SCAN \citep{lake2018generalizationsystematicitycompositionalskills} focuses on assessing whether the generated outputs exactly match the reference answers to indicate accuracy.
SyGuS \citep{DBLP:conf/iclr/OdenaSBSSD21} requires finding a program that satisfies the string transformation rule, and the number of tasks in which the correct program is successfully identified is counted as the success rate.

\subsubsection{Sandbox-based Evaluation}
\label{sec: sandbox}
Considering that all inductive reasoning tasks share the same intrinsic mechanism: inferring general rules from specific observations.
We can adopt a unified approach, namely sandbox unit test (Appendix~\ref{app: sandbox}), to standardize the evaluation across all the inductive benchmarks mentioned above.
It can be seen as a simplified agent system \citep{gan2026androidcoachimproveonline,chen2025guishepherdreliableprocessreward,su2026accuracyunveilinginefficiencypatterns}.

The sandbox unit test is a method where individual components or functions are tested in isolation to ensure they work as intended \citep{DBLP:conf/soups/AlhindiH25}. 
Each test runs in a controlled, independent environment, using specific input cases to verify the correctness of the component. 
This approach helps identify errors early and ensures that each part functions correctly before integration.

The sandbox unit test is originally used for code verification, at which time it is referred to as a code unit test \citep{DBLP:journals/corr/abs-2501-01277}.
With the development of LLMs, it is also applied to evaluate various deductive reasoning and agent-related tasks of LLMs, such as InternBootCamp \citep{DBLP:journals/corr/abs-2508-08636}.

\begin{figure}[t]      
  \centering        
  \includegraphics[width=1\linewidth]{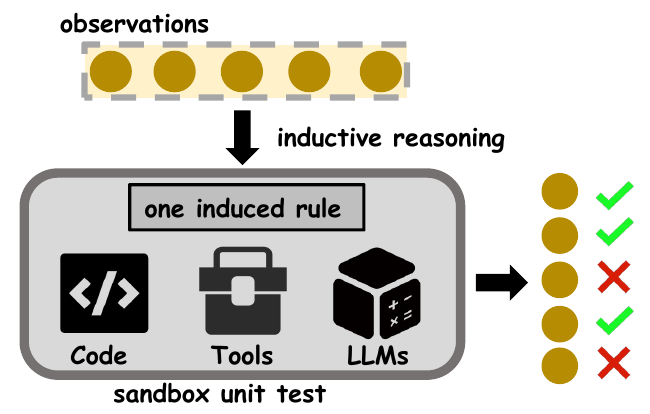} 
  \caption{A demo of the sandbox unit test for inductive reasoning of LLMs.}     
  \label{fig: sandbox}   
\end{figure}

In inductive reasoning tasks, the sandbox unit test can also be deployed for evaluation. 
We present a demo in Figure~\ref{fig: sandbox}. 
For an inductive rule generated by an LLM, it can be encapsulated as code \citep{DBLP:journals/corr/abs-2501-01277}, a tool \citep{DBLP:journals/fcsc/QuDWCWYXW25}, or written into a prompt to be provided to the LLM-as-a-Judge \citep{DBLP:journals/corr/abs-2411-15594}. 
Each observation can then be executed in a sandbox environment to determine whether it conforms to the current rule.

Based on this, we can derive a more fine-grained metric for LLM inductive reasoning: \textbf{observation coverage} (OC) (Appendix~\ref{app: oc}), defined as the proportion of observations that pass the unit tests out of the total number of observations.
In the example shown in Figure~\ref{fig: sandbox}, this value is $0.6$.
Compared with the overall ACC or success rate of a task, OC provides a more fine-grained supervision signal at the observation level. 
This allows for a more precise reflection of the comprehensiveness of the model’s answer. 
With this metric, more informative feedback can be provided for subsequent rule refinement and hypothesis exploration.

\section{Analysis}
\label{sec: analysis}

In this section, we present several prior exploratory tasks that offer theoretical analyses for inductive reasoning and inductive bias of LLMs \citep{DBLP:conf/iclr/KharitonovC21,DBLP:conf/emnlp/PapadimitriouJ23,wilson-frank-2023-inductive}.

\paragraph{Inductive ability originates from induction heads.}
Many studies \citep{DBLP:conf/acl/SiFJFC023,DBLP:conf/nips/Edelman0EMG24,DBLP:conf/nips/ChenSWY24} show that the strong in-context learning (ICL) \citep{DBLP:journals/corr/abs-2301-00234,DBLP:conf/emnlp/Dong0DZMLXX0C0S24,DBLP:conf/naacl/CrosbieS25} or example imitation \citep{DBLP:conf/acl/Honovich0BL23,DBLP:conf/acl/YeKK025} ability of LLMs originates from induction heads. 
An induction head is an attention head \citep{DBLP:conf/icml/EdelmanGKZ22,DBLP:conf/acl/RenG0LZQL24} that performs a match-and-copy operation, identifying and replicating relevant context tokens \citep{DBLP:conf/icml/SinghMHCS24}.
\citet{DBLP:journals/corr/abs-2505-16694} finds that, in fact, induction heads are meta-learning an abstract inductive within the context.

\paragraph{Model parameters, architecture, and data all help shape the inductive bias.}
The parameters, model architecture, and training data \citep{DBLP:conf/acl/WhiteC20, DBLP:conf/emnlp/MerrillRG0S21,DBLP:conf/iclr/LoveringJLP21,DBLP:conf/iclr/LevineWJNHS22,DBLP:conf/iclr/HaoChen023,DBLP:conf/iclr/MovahediOM25} are key to inductive bias.
\citet{DBLP:conf/nips/LipplL24} explores the effects of different parameters on inductive bias under multi-data mixed training and single-task finetuning scenarios, and ultimately emphasizes the importance of task similarity in mixed training.
Some studies \citep{DBLP:conf/icml/CabannesKBLB23,DBLP:conf/iclr/AerniMDY23} also highlight the importance of data augmentation, even the noisy data.
Further research \citep{DBLP:journals/corr/abs-2506-19031} demonstrates that the choice of minimum norm can also determine a model's inductive generalization.

\paragraph{Induction means simplicity.}

Some early studies show that complex model architectures and data \citep{DBLP:conf/icml/ZietlowRM21} can actually hinder inductive generalization.
At the same time, for higher-order models, regularization can actually be detrimental to the formation of their inductive bias \citep{DBLP:conf/iclr/PhuongL21,DBLP:journals/corr/abs-2203-03597}.
Sometimes, simplicity is perfect for inductive reasoning \citep{DBLP:conf/icml/GoldblumFRW24}.
To enhance the inductive reasoning ability, finding simple inductive bias is of paramount importance.
Simple and pure corpora often serve as the foundation for successful inductive reasoning \citep{DBLP:conf/acl/MuellerL23}.

\section{Conclusion}
This is the first survey of inductive reasoning for LLMs.
The inductive mode is crucial for knowledge generalization and aligns better with human cognition. 
We categorize methods for improving inductive reasoning into three areas: post-training enhancement, test-time exploration, and data augmentation. 
We also summarize the current benchmarks and derive a unified sandbox-based evaluation approach.
Finally, we offer some analyses regarding the source of inductive ability and how simple model architectures and data help with inductive tasks, providing a solid foundation for future research (Appendix~\ref{app: future}).

\section*{Limitations}
This paper is a survey about the inductive reasoning abilities of Large Language Models.
Due to space limitations, the main body of this survey is constrained to fewer than eight pages, and therefore, many details are not included in the main text. We only present the most essential parts. 
Meanwhile, although inductive reasoning in LLMs attracts increasing attention in recent years, the number of related studies remains relatively limited, making it difficult to produce a large-scale, systematic survey (even extending to $100$ pages) comparable to those in other areas.

\section*{Ethics Statements}
This survey primarily organizes and summarizes existing work on inductive reasoning in LLMs, with all relevant sources properly cited. Therefore, this paper does not raise any ethical or moral concerns.

\section*{Acknowledgments}
This work was supported in part by the National Natural Science Foundation of China (No. 62572198 and No. 92270119).
% We would like to especially thank Dezhao Ruan and Yuhao Dan for helping us read and organize a large number of papers.
% We also thank Linyang Li, Qipeng Guo, Prof. Xiaoming Shi, Wei Zhang, and other experts for their detailed suggestions.

\section*{Use of AI Assistants}
We primarily use AI assistants to improve and enrich our writing, especially by leveraging LLMs to help us write taxonomy in LaTeX.

%--------------------%
% 参考文献
%--------------------%
\bibliography{main}

%----------------------%
% 附录信息 (需要时可放开)
%----------------------%
% \appendix

% \section{Example Appendix}
% \label{sec:appendix}

% This is an appendix.
% \clearpage
\newpage
\appendix
\section{Appendix}

\subsection{Paper Collection}
Before writing our survey, we systematically search a large number of related papers and conduct a detailed categorization to include them in our GitHub repositories through two steps.

Step 1: Collecting inductive reasoning papers.
Since this work is in the field of NLP, we primarily focus on top-tier NLP conferences, including ACL, EMNLP, NAACL, COLING, and EACL, as well as the three major AI conferences—NeurIPS, ICML, and ICLR. 
Some additional papers from other venues, including those on arXiv, are also included.
The time range of our paper search spans $2015$ to $2025$, aiming to capture the major stages of the deep learning era.
We retrieve the accepted paper lists of each conference using keywords such as `induction', `inductive reasoning', `inductive study', and `inductive learning', etc.
We document papers containing the above keywords and upload most of them to our GitHub repository\footnote{https://github.com/141forever/inductive-reasoning-papers}. 
This allows us to gain a detailed understanding of the development of inductive reasoning.

Step 2: Focusing on LLM-related papers and refining the taxonomy.
To enhance the timeliness of our survey, we carefully select and read LLM-related papers, which serve as the basis for the initial draft of our survey taxonomy.
In addition, we sort papers on inductive reasoning for NLP downstream tasks in the PLM era by citation count, then read and summarize each one, abstracting their methods, and pick up the parts that overlap with approaches in the LLM era. We supplement them to further refine the taxonomy.
The final selected papers are provided in another repository\footnote{https://github.com/BDML-lab/llm-inductive-reasoning-survey}.

In summary, our paper selection criteria include (1) papers accepted by major AI conferences over the past decade, (2) all LLM-related and highly influential PLM-related inductive reasoning papers, (3) papers that propose a method, a dataset, or conduct some explorations. 
We exclude the papers that are: (1) without sufficient technical detail, (2) non-peer-reviewed blogs, (3) incomplete short papers, (4) low-cited papers that are not from the past two years, and so on.

\subsection{Inductive Reasoning and Inductive Bias}
The differences between inductive reasoning and inductive bias can be discussed in the following three aspects:

\paragraph{Different concepts} 
Inductive reasoning is a type of reasoning task whose input consists of some observations and whose output is a general rule or principle that is consistent with these observations. 
In contrast, inductive bias is a concept that represents the prior knowledge or inclination within a machine learning system \citep{DBLP:journals/corr/abs-2011-15091}.
It is the preferences or constraints that are `pre-specified' in the model before learning.
The process of learning such bias is called meta learning \citep{DBLP:journals/corr/abs-1810-03548, anonymous2026tamedit} in the machine learning field.

\paragraph{Different scopes of application} 
All machine learning and deep learning tasks (including text, vision, video, and other tasks) \citep{Dong2025InterleavedLV,Wang2025AutoregressiveSV,Li2025VideoProAP} inherently require the model or algorithm to be injected with the inductive bias appropriate for the specific task. 
Inductive reasoning also refers to one of the tasks.

\paragraph{Context of discussion} 
The goal of this inductive reasoning task is to identify rules among observations within a single task.
This paper primarily discusses the inductive reasoning task within the context of LLMs.
Improving the inductive reasoning ability of LLMs essentially amounts to discovering the inductive bias underlying inductive reasoning.
By doing so, the scope of the discussion can be narrowed, allowing the survey to be more thorough and well-developed.

\subsection{Inductive Reasoning and Other Reasoning Modes}
\label{app: diff}

\subsubsection{Inductive Reasoning and Deductive Reasoning}
\begin{figure}[h!]
    \centering
    \includegraphics[width=0.5\textwidth]{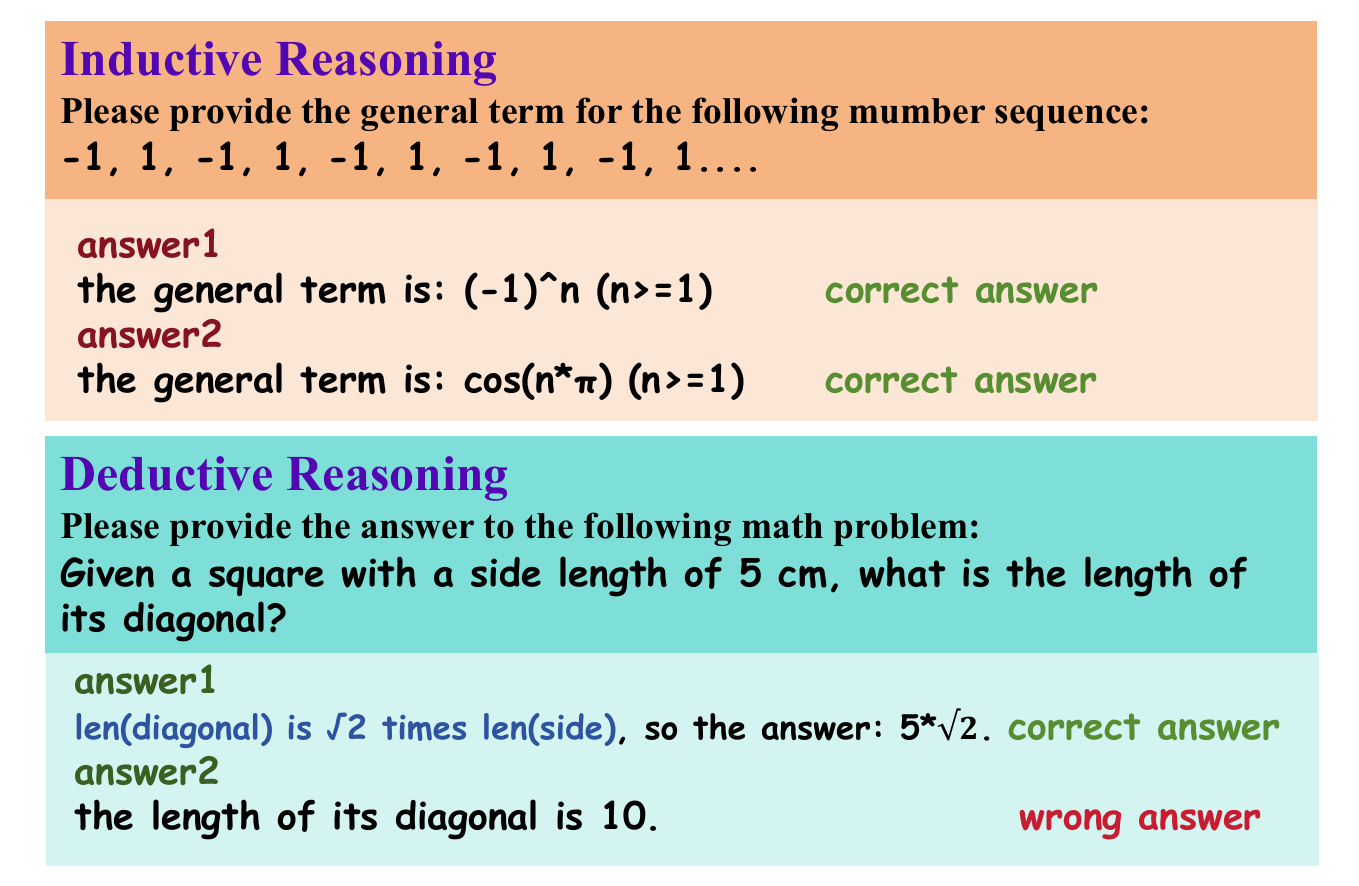} % 图片文件名
    \caption{An example is given for both reasoning modes. In inductive reasoning, there may be multiple correct answers consistent with the existing observations, while in deductive reasoning, a precise logical reasoning process can help arrive at the only correct answer.}
    \label{fig:cases}
\end{figure}

We provide an example for each in Figure~\ref{fig:cases}.
As two major modes of reasoning, inductive reasoning \citep{Clark1969LinguisticPI, Rips1994ThePO, Bang2023AMM} differs from deductive reasoning in many aspects.
Please refer to Table~\ref{tab: diff} for more.

\begin{table}[h!]
\centering
\renewcommand{\arraystretch}{1.2}
\caption{The differences between inductive reasoning and deductive reasoning.}
\resizebox{1\linewidth}{!}{
\begin{tabular}{ccc}
\toprule
 & \textbf{Inductive Reasoning} & \textbf{Deductive Reasoning} \\ 
\midrule
\textbf{Mode} & particular-to-general & general-to-particular \\
\textbf{Target} & probabilistic conclusion & precise answer \\
\textbf{Reliability} & flexible and generative & premises-relied \\
\textbf{Informative} & extended information & no new information  \\
\textbf{Application} & hypothetical inference & rigorous proof \\
\bottomrule
\end{tabular}
}
\label{tab: diff}
\vspace{-0.5em}
\end{table}

\subsubsection{Inductive Reasoning and Analogical Reasoning}
As stated in the main text, inductive reasoning is a process that goes from the particular to the general. 
In contrast, analogical reasoning is a process from the particular to the particular, which can be regarded as a special form of inductive reasoning. 
For example, consider an inductive reasoning task of deriving the general term formula of a number sequence: the input is a number sequence, and the output is its general term formula. 
Analogical reasoning, on the other hand, takes a number sequence as input and outputs its next term. 
In short, analogical reasoning is a form of reasoning that involves imitation based on observations. 
It compares and transfers similarities from existing observations to infer and generate possible next items or outcomes \citep{lewis2024evaluating,lewis2024using,musker2024semantic}.

In the field of LLMs, the commonly studied ICL \citep{DBLP:journals/corr/abs-2301-00234,DBLP:conf/emnlp/Dong0DZMLXX0C0S24} can be regarded as a form of analogical reasoning. 
For ease of discussion and to clearly define the scope, a large body of research on ICL is not discussed in this paper.

\subsection{More Synthesis and Comparative Analysis on NLP Applications}
\label{app: nlp}

The impressive applications we cited, spanning from core syntactic and semantic parsing to complex multimodal tasks, clearly demonstrate the versatility of inductive reasoning in NLP.

\textbf{Empirical effectiveness} often hinges on the method's ability to seamlessly integrate domain-specific structural knowledge as inductive bias — for instance, achieving high precision in information extraction usually involves explicit rule-based or graph-structured inductive biases, while improvements in dialogue systems frequently rely on incorporating conversational history as an inductive structure.

Approaches that embed strong, hand-crafted inductive bias (e.g., explicit grammar rules for parsing) tend to offer \textbf{superior interpretability and data efficiency} but suffer from lower generalizability across domains, whereas soft or learned bias (e.g., the knowledge encoded in the model’s parameters) is \textbf{more flexible but potentially less transparent}. 
How to achieve a \textbf{trade-off} between the two is currently a key focus for applying inductive reasoning to downstream NLP tasks.

\textbf{The major trend} of inductive reasoning in NLP is a move toward methods that automate the discovery of inductive patterns, such as applying self-supervision or meta-learning to identify bias suitable for multiple tasks, marking a shift from prescriptive to descriptive induction \citep{Kalish2014DescriptiveAI} in method design. 

This comparative analysis underscores the necessity of selecting an inductive reasoning method that carefully balances the need for robustness, interpretability, and generalization capacity for the target application scenario.

\subsection{Inductive Reasoning Across Different Eras}
In the traditional deep learning or PLM era, inductive reasoning was mostly treated as a core idea for addressing downstream NLP tasks (Section~\ref{sec:app-nlp}), such as designing inductive network architectures or introducing inductive\/analogical data.
Although some benchmarks directly targeting inductive reasoning were proposed (upper part of Table~\ref{tab: bench}), they were largely used to study neuro-symbolic reasoning or meta-learning.

The LLM era has developed rapidly, and the generative paradigm has gradually unified most downstream NLP tasks. 
As the scale of data and model parameters continues to grow, many issues that traditional meta-learning focused on—such as catastrophic forgetting and domain adaptation—have been greatly alleviated. 
Consequently, with the increasing foundational capabilities of models, inductive reasoning has emerged as a standalone task for research. 
This paper focuses primarily on this task in the LLM era, providing a comprehensive literature review and discussion.

In the methodology discussion of Section~\ref{sec: method}, Sections~\ref{sec: post-train} Post-training Enhancement and Section~\ref{sec: hypo} Test-time Exploration are concepts and methods unique to the LLM era, whereas some sub-methods in Section~\ref{sec: info} Data Augmentation are derived from traditional deep learning approaches to inductive reasoning, such as introducing additional inductive-enhanced data.

\subsection{The Boundaries and Comparisons among the Methods}
\label{app: med}

\subsubsection{The Boundaries}
Our classification minimizes overlap among the three methods conceptually as much as possible.

(1) \textbf{Post-training enhancement} directly trains LLMs using task-specific data. 
Synthetic data in this category is often generated from scratch to create entirely new training samples.
(2) \textbf{Test-time exploration} is a train-free method based on inductive hypotheses.
(3) \textbf{Data augmentation} can guide inductive tasks by leveraging annotated data and knowledge from outside the task domain manually. 
It can keep the model frozen, or even be used to train with fine-grained embedding or graph information.
In conclusion, it modifies and expands existing training samples.

We believe that modern LLM techniques are often multifunctional, and some of the overlapping parts can be seen as emerging trends or directions for future research.
For example, the construction of synthetic data inevitably incorporates prior knowledge from human experts.

\subsubsection{The Comparisons}

(1) \textbf{Post-training enhancement} is suitable for scenarios where sufficient data is available, and there is a need to directly enhance a single model’s inherent inductive capabilities. 
\textit{Advantages}: It can directly enhance the inductive reasoning ability of a specific LLM and is easy and quick to train.
\textit{Disadvantages}: It is only applicable to scenarios involving a single model, and cannot be easily used to solve complex problems that require cooperation among agent systems.

(2) \textbf{Test-time exploration} can be applied in situations with limited data or when the model cannot be fine-tuned, using hypothesis-driven methods to perform inductive reasoning tasks.
\textit{Advantages}: No model training is required, as it can leverage the capabilities of powerful closed-source LLMs.
\textit{Disadvantages}: It needs time and monetary costs, and multiple refinements do not necessarily guarantee an optimal solution.

(3) \textbf{Data augmentation} is more appropriate for scenarios where the model involves cross-domain tasks or requires the use of external knowledge (e.g., cross-domain agent systems with tool usage and human-crafted rules), or for training auxiliary modules such as embeddings or graphs, thereby improving the model’s generalization and transferability to new patterns.
\textit{Advantages}: It can leverage information beyond the LLM, including even parameter-level information, to provide assistance.
\textit{Disadvantages}: Its working mode is similar to that of an agent system, giving it potential to handle complex real-world problems. However, it relies on external data and tools, and the overall system performance depends on the fundamental capabilities of each component.

\subsubsection{Why Synthetic Data is Widely Used}
Synthetic data accounts for a significant portion of many well-known LLM training corpora \citep{DBLP:journals/corr/abs-2412-19437,DBLP:journals/corr/abs-2505-09388,DBLP:journals/corr/abs-2508-15763} and has the following advantages compared with real data.

(1) \textbf{Coverage of rare cases}: Synthetic data can include samples that are uncommon in real-world scenarios. For example, in a batch of real data, certain inductive patterns may appear infrequently, and directly training on such data could introduce bias into the model.
(2) \textbf{Support for data-scarce or privacy-sensitive domains}: In fields where data is limited or highly sensitive, synthetic data can be generated at a large scale, enabling the training of more generalizable models.
(3) \textbf{High efficiency}: Synthetic data can be generated very flexibly and at a large scale, which makes model training and debugging more efficient.
(4) \textbf{Sometimes better performance}: In certain cases \citep{DBLP:conf/iclr/YuanZS0Z24,DBLP:journals/corr/abs-2510-08095}, synthetic data can even yield better results. Because it may better fit the model’s distribution and even provide regularization.

For these reasons, it is also a better choice for inductive reasoning.

\subsection{More Details about Benchmarks}
\label{app: bench}

In this section, we will introduce the inputs and outputs of LLM inductive reasoning benchmarks one by one in detail.

\paragraph{ILP}\citep{evans2018learningexplanatoryrulesnoisy}. It takes as input background knowledge in first-order logic, along with a pair of positive and negative examples for a specific first-order logic case, and the model is required to output a first-order logic that satisfies these conditions.

\paragraph{SCAN}\citep{lake2018generalizationsystematicitycompositionalskills}. It takes as input a series of entities and their states, and requires LLMs to output the actions needed to achieve those states. 
For example, given `a ball on the table' as input, the model should output `place it on the table'.

\paragraph{ARC}\citep{DBLP:journals/corr/abs-1911-01547}. It takes as input several pairs of grids in natural language form, where they illustrate a specific transformation pattern. 
Then, given a new grid as input, the model is asked to output what the transformed grid would look like.

\paragraph{List Functions}\citep{rule2020child}. It takes as input several pairs of number lists, where they illustrate a specific transformation pattern. 
Then, given a new number list as input, the model is asked to output what the transformed number list would look like.

\paragraph{PROGES}\citep{DBLP:conf/icml/AletLKNSLKT21}. It provides several input-output pairs of a program and requires the model to generate the program itself.

\paragraph{SyGuS}\citep{DBLP:conf/iclr/OdenaSBSSD21}. It takes as input several pairs of strings, where they illustrate a specific transformation pattern. 
Then, it requires the model to generate a program to show this transformation process.

\paragraph{ACRE}\citep{DBLP:conf/cvpr/0017JEZZ21}. It takes as input the results of interactions between different entities and a certain machine (i.e., the functions of different entities), and models need to output the entity corresponding to a specific function.

\paragraph{Instructions}\citep{honovich2022instructioninductionexamplesnatural}. The model is given two natural language statements, A and B, where B is obtained by applying a certain instruction to A, and it is required to output that natural language instruction.

\paragraph{Arithmetics}\citep{DBLP:conf/naacl/WuQRA0WKAK24}. It takes as input several pairs of two-digit additions along with their results, where these additions follow a calculation process in a certain numeral base. 
Given a pair of two-digit numbers, the model is required to output the result of their addition in the same base.

\paragraph{Levy/Holt}\citep{DBLP:conf/emnlp/LiuLCS24}. It takes as input a pair of triplets, where the positive triplet represents a factual relationship and the negative triplet represents a counterfactual relationship. 
The task is to output an inference rule between triplets such that the positive triplet entails the negative triplet.

\paragraph{NutFrame}\citep{DBLP:conf/coling/Guo000024}. It takes text fragments that contain potential frames and frame elements, along with contextual information such as sentence structure, lexical cues, and semantic hints. This input helps the model identify underlying conceptual structures in the text.
The model produces three types of outputs.
Frame Induction: Identifies latent frames expressed in the text and maps them to existing FrameNet frames or proposes new frames.
Frame Element Identification: Detects specific frame elements within the text.
Frame Filling: Assigns concrete values from the text (entities or phrases) to the identified frame elements.

\paragraph{DEER}\citep{DBLP:conf/eacl/YangDDCCLGW24}. It takes as input several pairs of facts, where they illustrate a specific real-world rule. 
Then, it requires the model to generate the rule.

\paragraph{RULEARN}\citep{he2025ideaenhancingrulelearning}  The input to it consists of three parts.
Puzzle Scenarios: Each puzzle presents a set of conditions or operations that the agent can manipulate. 
These scenarios are designed to have underlying rules that are not explicitly provided.
Agent Actions: The agent can perform a variety of actions, such as inputting integers or letters, to interact with the puzzle environment.
Feedback: After each action, the agent receives feedback that helps in refining its understanding of the hidden rule.
The desired output is: the hidden rule governing the puzzle scenario based on the feedback received from its actions.

\paragraph{Cryptography}\citep{DBLP:conf/acl/Li0ZS25} It takes as input several pairs of English words, where each pair follows a certain cryptographic transformation pattern.
Given a new word, the model is required to output the new word obtained by applying the same transformation pattern.

\paragraph{GeoILP}\citep{DBLP:conf/iclr/ChenZZ25} The same as ILP in general.

\paragraph{InductionBench}\citep{DBLP:conf/acl/HuaWSPJW25} It takes as input a pair of strings, where the pair follows a certain transformation rule, and the task is to output that rule.

\paragraph{CodeSeq}\citep{chen2025codedriveninductivesynthesisenhancing} It takes as input a number sequence of numbers and is required to output the number sequence’s general formula, with the entire output presented in code form.

\textbf{Statements} We determine the number of test samples for each benchmark by consulting the original paper, the corresponding GitHub repository, and the Hugging Face dataset files, either through direct counting or estimation.

\subsection{Failure Analysis}
\label{app: fail}
Different benchmarks evaluate different models and do not provide a standardized, unified evaluation strategy. 
Therefore, for the sake of fairness, we cannot provide directly comparative results.

Hence, we summarize several typical failure modes of LLMs on these inductive tasks based on prior experiences and the conclusions drawn in previous studies.

(1) \textbf{Difficulty in internalizing inductive logic.}
LLMs tend to capture only surface-level associations among observations, while failing to truly acquire the underlying logical or mathematical relations. 
For example, when presented with new samples or longer samples following the same pattern, the model often cannot induce the correct rule or produce the correct answer.
(2) \textbf{Spurious pattern matching.}
LLMs often latch onto accidental or superficial patterns that happen to fit the limited observations, leading to spurious rules that do not reflect the true underlying structure. 
These rules typically fail on even slightly varied test cases, revealing that the induced pattern was an artifact rather than a genuine abstraction.
(3) \textbf{Failure in multi-step or compositional reasoning.}
Many inductive tasks require composing several primitive operations (e.g., nested arithmetic, layered string transformations, or multi-step grid manipulations). 
LLMs often fail to correctly chain these operations, producing partial reasoning steps that appear correct individually but do not combine into a coherent global rule.
(4) \textbf{Syntactic formatting errors.}
The rule induced by the model may be semantically correct, but it does not conform to the syntactic constraints of the task or the underlying logical system. 
For example, the generated code-like rule may contain syntax errors, making it impossible to execute.
(5) \textbf{Data sparsity challenge.} 
The rules induced by the model often apply only to the most frequent observations. 
When the distribution of inductive patterns is uneven, the model is prone to errors on rare or boundary-case observations that lie in the long tail.

\subsection{Advantages of Sandbox and OC}

\subsubsection{Advantages of Sandbox}
\label{app: sandbox}
We will elaborate on the advantages of sandbox testing from the following points.

(1) \textbf{Meeting the demand for complex real-world tasks}: 
With the development of LLMs, there is an increasing demand for handling more complex tasks in real-world scenarios, such as scientific discovery and simulations. 
Consequently, Agent technologies have also advanced rapidly. 
In these scenarios, task success rates cannot be adequately measured using traditional metrics. 
Sandbox testing, as a highly integrated evaluation framework combining LLMs, tool usage, and complex rules, is better suited to address modern AI tasks, including inductive-related tasks.
(2) \textbf{Isolation and safety}: 
Sandboxes are typically isolated environments with higher security, which helps reduce bias and risk while allowing rapid problem detection. 
In contrast, traditional metrics provide only coarse numerical indicators.
(3) \textbf{Scalability}: 
Sandboxes are scalable, and existing sandbox frameworks can cover thousands of LLM tasks \citep{DBLP:journals/corr/abs-2508-08636}.
(4) \textbf{Many works adopt a sandbox as the evaluation tool}: Their experimental results demonstrate that a sandbox can reveal a greater variety of errors \citep{DBLP:journals/corr/abs-2503-04479}, provide more logical explainability \citep{DBLP:journals/corr/abs-2509-20364}, and cover more boundary behaviors \citep{Alshahwan2024AutomatedUT}.

Many existing LLM foundation trainings, such as the Intern series, as well as the Minimax-M1 and Seed-thinking models, employ large amounts of all kinds of reasoning data and sandbox validation to enhance the models’ fundamental reasoning capabilities.

\subsubsection{Advantages of OC}
\label{app: oc}

The OC evaluation method has the following advantages.

(1) \textbf{Overcoming averaging effects}:
Metrics like ACC only measure overall accuracy, reflecting the average performance, but they do not reveal whether the model works effectively across all observations. 
Measuring OC provides a more fine-grained evaluation signal.
(2) \textbf{Improved explainability}: 
This observation-level evaluation can uncover a greater variety of error types, allowing analysis of specific errors.
(3) \textbf{Empirical support from existing work}: 
Previous studies \citep{chen2025codedriveninductivesynthesisenhancing} (it reformulates number sequence as algorithmic problems and uses the OC of code solutions as the training signal) show that using observation-level supervision signals can effectively enhance a model’s inductive reasoning abilities.

\subsection{Future Works}
\label{app: future}

We will further expand and discuss the prospects and potential applications of inductive reasoning in this subsection.
\subsubsection{In AI domain}

\paragraph{Inductive reasoning tasks, benchmarks, and data} 
The inductive reasoning capability of LLMs can be examined from various other perspectives, such as format imitation \citep{DBLP:journals/corr/abs-2408-00114}, cross-domain induction \citep{DBLP:conf/eacl/ChenSM24}, multimodal inductive \citep{DBLP:conf/iclr/WangZPPHG24}, and so on.
Even a single data point has the potential to be expanded into either SFT or RL data, like some other works \citep{guo2026rethinkingmultiplechoicequestionsrlvr}. 
This serves as a concrete truth demonstrating the scalability of inductive reasoning data. 
Therefore, constructing more inductive reasoning datasets and evaluation methods is key to improving LLMs’ inductive reasoning capabilities.

\paragraph{Enhancing the inductive reasoning ability of LLMs} 
We summarize several typical failure modes of LLMs on inductive tasks in Appendix~\ref{app: fail}. 
Addressing these issues can further enhance the model’s inductive reasoning ability. 
Using simple and effective training data and architectures to identify universal inductive biases is a reasonable option.
In Section~\ref{sec: analysis}, we conclude some analyses of inductive reasoning theoretically from current practical experience.
These analyses are highly beneficial for improving the inductive reasoning capabilities of LLMs.
The paper also presents three major categories of methods, and it is expected that more methods will be proposed in the future.

In particular, future work should focus on three practical directions. 
First, since models often rely on superficial correlations rather than internalizing rules, a feasible direction is to construct controlled synthetic training sets that isolate the true inductive relation. 
Concretely, one can generate training pairs that preserve the underlying rule while varying surface patterns and apply balanced sampling across pattern types. 
Second, to address failures in compositional induction, models can be trained with explicit intermediate representations and verification loops. 
For example, program-of-thought or code-based reasoning can externalize rule composition into executable steps, while multi-round reasoning or critic-guided refinement frameworks allow the model to iteratively revise intermediate rules. 
Third, because current models are sensitive to long-tail and boundary cases, evaluation and training protocols should include distributional perturbations. 
This can be implemented by augmenting data with rare-pattern samples, perturbing sequence length or structure, or even adding contrast sets that break spurious heuristics.

Considering that PBE is the longest-established inductive reasoning task with the richest body of research, we will list several algorithms related to solving the PBE task below.

\textbf{Example I: PBE}
Inductive program synthesis (or programming‑by‑example, PBE) is fundamentally an inductive reasoning task and one of the earliest to originate: given a few procedural input-output pairs, the system must generalize to unseen inputs.
\citet{DBLP:conf/nips/LiE24} argues that PBE is a highly general form of few-shot inductive inference, and they show that while pretrained LLMs perform poorly on classic PBE tasks (e.g., list and string manipulations), fine‑tuning on in‑distribution PBE data substantially improves performance.
To address the inherent lack of search in LLMs, \citet{DBLP:conf/iclr/Verbruggen0S0G25} introduce a within‑prompt search method: they sample lines of code, execute them on the input examples, feed back the semantic outcomes, and use that to guide further generation, effectively letting the model perform a semantic‑aware search.
Building on this inductive foundation, \citet{DBLP:journals/corr/abs-2509-17393} propose a transductive framework: at test time, they actively select test inputs, have the LLM predict their outputs, and eliminate candidate programs that disagree — this refines the inductive hypothesis space using feedback.
In addition, recent work such as \citet{DBLP:journals/corr/abs-2503-15540} breaks down complex PBE tasks into subtasks, with the LLM guiding the synthesis of these subtasks and then recombining them, thus improving the model’s ability to solve difficult problems.
At the same time, the CodeARC \citep{DBLP:journals/corr/abs-2503-23145} benchmark has been introduced to evaluate LLMs’ reasoning ability in real-world inductive synthesis settings: the model can interactively query a black-box target function, propose candidate functions, and iteratively refine them via a differential testing oracle.

These advances collectively show that under LLM-driven paradigms, PBE is being reactivated and reshaped: modern methods integrate fine-tuning, search, semantic execution, and increasingly also active learning and feedback mechanisms to boost generalization and robustness.

\subsubsection{In Real-world Domain}
\paragraph{Driving scientific discovery and innovation} 
In the context of scientific research, inductive reasoning is a vital method for uncovering natural laws. 
Artificial intelligence, especially generative AI, is becoming a new tool for scientists to propose hypotheses and design inductive experiments \citep{DBLP:conf/aaai/ReddyS25}.
By learning patterns from large-scale data, these models can suggest plausible explanations and explore vast hypothesis spaces that would be difficult for humans to enumerate manually.
This can serve a wide range of scientific fields, including the technological applications mentioned in Section~\ref{sec:app-real}.

Since LLMs often fail for superficial correlations, we will highlight that AI-assisted scientific discovery should incorporate iterative hypothesis testing, simulation, or tool-based verification. 
This helps ensure that induced rules reflect underlying mechanisms rather than accidental patterns.

\textbf{Example II: AI4Science}
AI4Science constitutes a form of inductive reasoning by LLMs: instead of merely fitting data, LLMs generate symbolic hypotheses (programs) that generalize from observed examples, thereby performing induction over scientific phenomena.
For instance, \citet{DBLP:conf/iclr/ShojaeeMGFR25} proposes new equation `skeletons' using an LLM informed by scientific priors, and refines them with evolutionary search to discover mathematically and physically meaningful relations.
This work touches on debates about the nature of scientific reasoning.
\citet{Barnum2012TheBO} argues that science is not purely induction, which raises interesting questions about whether LLM‑based equation discovery should be framed as induction, deduction, or something more agentic.
More broadly, \citet{DBLP:conf/acl/MerlerHDM24} uses LLMs to iteratively propose functions and optimize coefficients, reinforcing that LLMs are now capable of hypothesis generation rather than mere curve-fitting.
Extensions like \citet{DBLP:journals/corr/abs-2506-04282} further enrich this paradigm by combining data-driven insights with a reflective feedback loop for refining symbolic candidates.
Taken together, these developments suggest that LLM-driven scientific discovery can be viewed as one of the future works of inductive reasoning.

\paragraph{Human–machine collaboration in education} 
As cognitively‑enhanced AI develops, future educational systems may evolve into "human–in-the-loop" intelligent agents \citep{DBLP:journals/corr/abs-2503-11733}. 
Inductive reasoning can help AI systems better understand students’ ways of thinking and their learning processes.
By inferring latent patterns from students’ interactions and responses, such systems can adapt instruction strategies to individual needs and learning trajectories. 
Furthermore, inductive reasoning enables AI to provide more interpretable feedback and personalized guidance, fostering more effective and engaging human–AI collaboration in education.

In educational settings, one can induce students’ fundamental cognitive and thinking patterns based on their behavior; however, these patterns are composite and therefore highly complex. 
As a result, instructor feedback can dynamically adjust the framework for interpreting student behavior, enabling the system to refine its inductive hypotheses and better align with individual learning processes.

\paragraph{New perspectives from philosophy and cognitive science}
From the perspective of cognitive science, inductive reasoning is a crucial tool for understanding human thought and the structure of cognition. Future research may further explore the similarities and differences between the inductive reasoning of AI and that of human thinking \citep{Trepczyski2024ReligijaTI}.
Such investigations could shed light on whether AI systems rely on mechanisms analogous to human abstraction, generalization, and concept formation, or whether they achieve inductive competence through fundamentally different processes.

Observing failures such as syntactic inconsistencies or difficulties with sparse data in AI models highlights aspects of human cognition that are sensitive to rule structure and evidence distribution. Such comparisons can inspire theories about how humans integrate formal constraints and limited observations to achieve robust inductive competence, deepening our understanding of intelligence and reasoning.

\paragraph{Ethical and social implications} 
As inductive reasoning capabilities become more powerful, AI systems’ decisions or recommendations may embody inductive generalizations that `seem reasonable but could be wrong'. Avoiding inductive bias will thus become a significant issue \citep{Ueno2022TrustIH}.
Such biases may arise from spurious correlations, incomplete data, or mismatches between training environments and real-world contexts, leading to systematic errors that are difficult to detect. 
Addressing this challenge will require not only improved model design and evaluation but also greater transparency, uncertainty awareness, and human oversight to ensure that inductive inferences remain reliable and trustworthy.

Therefore, the continual evolution of inductive reasoning is one of the key approaches to addressing the common error patterns identified in the Appendix~\ref{app: fail}, and even to advancing societal ethics and moral standards.

\end{document}